\documentclass[11pt]{article}

\usepackage[preprint]{acl}

\usepackage{times}
\usepackage{latexsym}

\usepackage[T1]{fontenc}

\usepackage[utf8]{inputenc}

\usepackage{microtype}

\usepackage{inconsolata}

\usepackage{graphicx}
\usepackage{booktabs}
\usepackage[table]{xcolor}

\usepackage{amsmath,amssymb}

\usepackage{marvosym}

\title{Which Pretraining Paradigm Better Serves Spatial Intelligence? \\ An Empirical Comparison of Vision-Language \\ and Video Generation Models}

\author{%
    \textbf{Haozhan Shen$^{1}$} \quad
    \textbf{Tiancheng Zhao$^{2,3}$\textsuperscript{\Letter}} \quad
    \textbf{Kangjia Zhao$^{1}$} \quad
    \textbf{Jianwei Yin$^{1}$}\\[3pt]
    $^1$ Zhejiang University \quad
    $^2$ Om AI Research \quad
    $^3$ Binjiang Institute of Zhejiang University \\
    \small{
       \Letter\ Correspondence: {tianchez@zju-bj.com}
     }
    \vspace{-3mm}
}

\begin{document}
\maketitle
\begin{abstract}
Spatial intelligence requires visual representations that capture both semantic objects and geometric structure in the physical world.
To support this, two major pre-training schemes are now widely used as foundation backbones: Vision-Language Models (VLMs), which use language supervision to align visual observations with semantic concepts, and Video Generation Models (VGMs), which learn from temporally evolving visual worlds.
However, it still remains unclear which pre-training scheme provides a better representation substrate for spatial intelligence.
In this paper, we present the first systematic frozen-feature probing study of VLMs and VGMs across three representative axes of spatial intelligence: semantic tagging, instance grouping, and 3D geometry prediction.
Using the lightweight probe, our framework enables a controlled comparison of what information is already encoded in frozen representations from two model families.
Experimental results reveal a clear complementarity: VLMs are stronger at semantic tagging and instance grouping, while VGMs provide more accessible signals for dense geometry and camera motion.
Moreover, a naive fusion of the two already yields a representation that excels at both geometry and semantics, suggesting a promising direction for building stronger spatial-intelligence backbones by effectively integrating features from both model families. 
Our code is available at \href{https://github.com/om-ai-lab/Probing-VLM-VGM}{https://github.com/om-ai-lab/Probing-VLM-VGM}.
\end{abstract}

\section{Introduction}

\begin{figure}[t]
    \centering
    \includegraphics[width=\columnwidth]{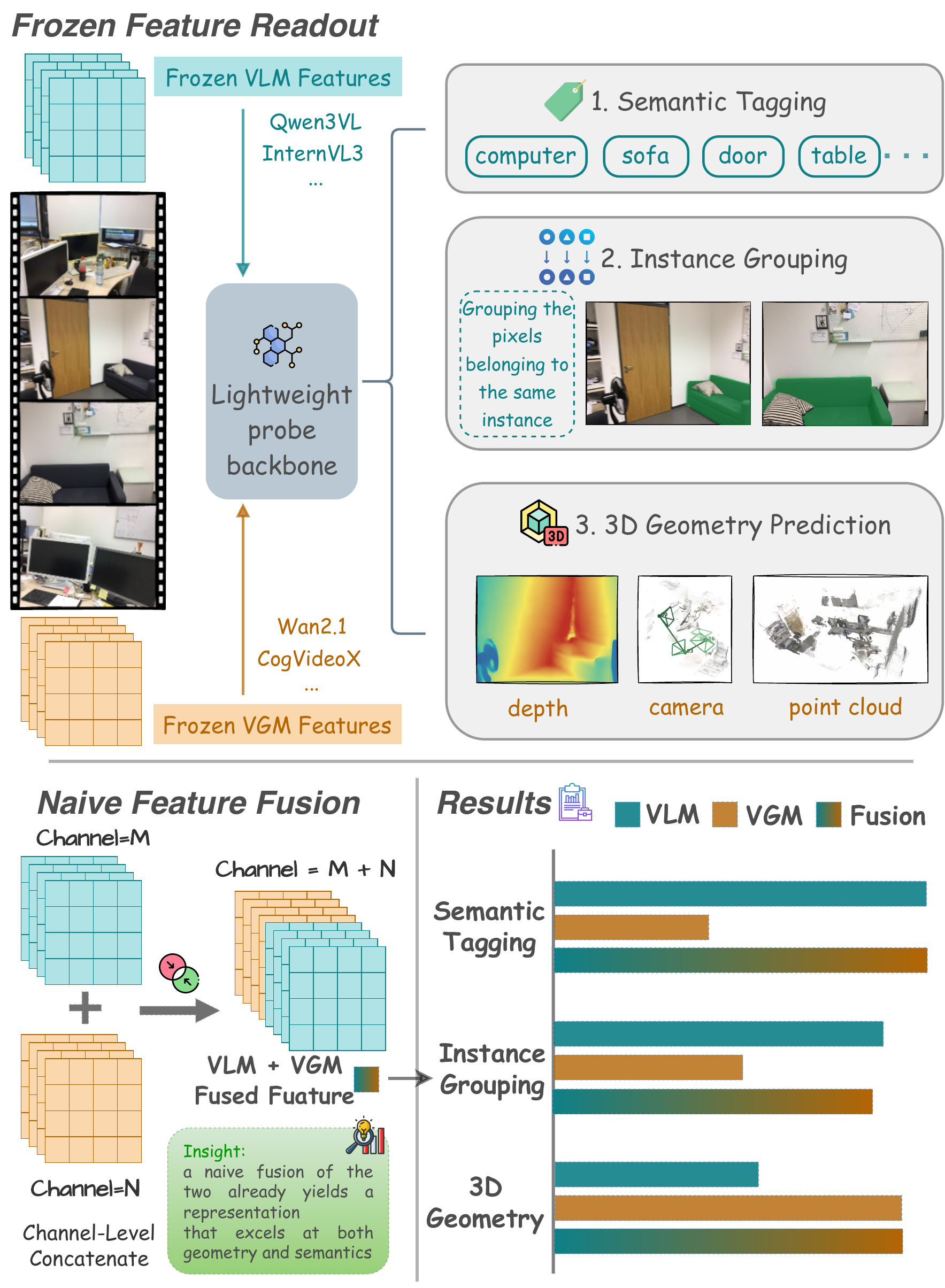}
    \vspace{-4mm}
    \caption{
        \textbf{Top:} Frozen VLM and VGM features are probed on three axes for spatial intelligence.
        \textbf{Bottom:} Results show that VLMs excel at semantics and instances, VGMs excel at geometry, and simple fusion combines their strengths, further suggesting that the two representation families are complementary.
        }
    \label{fig:vlm-vgm-overview}
    \vspace{-4mm}
\end{figure}

Spatial intelligence is becoming an increasingly important research focus, as it is central to embodied AI~\cite{wang2024embodiedscan,majumdar2024openeqa,zhu2024spa,du2024embspatialbench}, autonomous driving~\cite{li2024bevformer,sima2024drivelm,gao2025survey,zhou2025opendrivevla}, real-world scene understanding~\cite{chen2024spatialvlm,li20243dmit,pan2025metaspatial,baruch2021arkitscenes}, and robotics~\cite{lin2024bip3d,ma2026spatialreasoner,qiao2025navbench,hughes2022hydra}, all of which are essential for building AI systems that can operate in and assist with diverse aspects of the physical world. 
Recent spatial-intelligence-specific training and evaluation efforts further reflect this trend~\cite{sensenova-si,yang2025thinking}.

In these settings, intelligent systems are expected to recognize relevant objects, understand scene layouts, and reason about spatial relationships for navigation, interaction, planning, and decision making~\cite{chaplot2020object,das2018embodied}.
Thus, the adopted visual representations must not only capture semantic object information, but also preserve the geometric structure.
To build such representations, recent work has largely followed two major pre-training schemes that are widely used as visual backbones:
Vision-Language Models (VLMs)~\cite{bai2025qwen3-VL,bai2025qwen2.5-VL,zhu2025internvl3,coreteam2025mimovltechnicalreport,vteam2025glm45vglm41vthinkingversatilemultimodal,liu2023visual} use language supervision to align visual observations with semantic concepts, while Video Generation Models (VGMs)~\cite{wan2025wan,zheng2025open-sora,zhu2025aether,yang2025cogvideox,genmo2024mochi,kong2024hunyuanvideo} learn from temporally evolving visual worlds.
The growing use of both families as perception backbones raises a central question: \textbf{which pre-training scheme provides a better representation, or do they excel at different aspects?}

Some recent work has begun to place these two routes within a common embodied-AI framework.
On one hand, modular VLA frameworks~\cite{community2026starvla} support both VLM and VGM backbones under a shared action-learning interface.
On the other hand, some World Action Model studies~\cite{zhang2026world,ye2026worldactionmodelszeroshot,team2026motubrain} compare video-based agents with VLA policies. 
However, they mainly evaluate complete policies rather than the pre-trained visual representations themselves, leading to a key limitation:
the observed performance is entangled with many factors, such as action decoders, robot data mixtures, post-training recipes, inference strategies, and benchmark protocols.
As a result, such comparisons cannot isolate the effect of the pre-training scheme, leaving unclear what spatial information is already encoded in the frozen representations of VLMs and VGMs before any downstream fine-tuning.

In this paper, we address this question through a controlled frozen-feature probing study of VLMs and VGMs for spatial intelligence.
Instead of fine-tuning the foundation models themselves, we freeze each model, extract visual features from its intermediate layers, and train lightweight probes with an identical backbone architecture and task-specific readout heads.
This design allows us to compare what information is readily recoverable from each pre-training scheme, while reducing confounding factors from model-specific decoders and downstream fine-tuning.

We operationalize spatial intelligence along three complementary probing axes
First, \emph{semantic tagging} asks which object categories appear in a video, capturing the semantic object knowledge needed for instruction grounding, task planning, and object discovery.
Second, \emph{instance grouping} asks which pixels across multiple views belong to the same object instance, reflecting object-centric representations needed for object permanence, tracking, and interaction.
Third, \emph{3D geometry prediction} asks the probe to recover dense depth, point maps, and camera motion, measuring the geometric structure needed for scene layout understanding, navigation, and physical reasoning.
Together, these tasks span a spectrum from recognizing \emph{semantic objects} to reconstructing \emph{spatial worlds}.

Concretely, we use lightweight probes to read out frozen features from representative VLMs such as Qwen3-VL~\cite{bai2025qwen3-VL} and InternVL3~\cite{zhu2025internvl3}, and VGMs such as WAN2.1~\cite{wan2025wan} and CogVideoX~\cite{yang2025cogvideox}, along these three axes, as shown in Figure~\ref{fig:vlm-vgm-overview} Top.
Notably, our experiments reveal a clear complementarity rather than a single winner.
VLMs consistently outperform VGMs on semantic tagging, showing that language-aligned pre-training makes object-category information more accessible.
Moreover, VLMs also perform better on instance grouping, suggesting that grouping object instances across views benefits strongly from object-centric semantic representations, not only from geometric continuity.
In contrast, VGMs achieve stronger results on 3D geometry prediction, where dense point maps, depth, and camera motion are more accurately recovered from video-generation features.
This pattern suggests that current VLMs are better at encoding semantic objects, whereas current VGMs are better at encoding spatial geometry.

This complementarity also has a constructive implication: as shown in Figure~\ref{fig:vlm-vgm-overview} Bottom and §\ref{sec:fusion}, a simple feature-level fusion of normalized VLM and VGM features already preserves the VLM semantic advantage while recovering much of the VGM geometric strength.
This suggests a path toward spatial-understanding backbones that integrate language-aligned object representations with generation-induced geometric representations.

Our contributions are summarized as follows.
\textbf{(1)} We formulate a frozen-representation comparison between VLMs and VGMs for spatial intelligence before downstream fine-tuning, avoiding confounds from action heads, robot data, and post-training recipes.
\textbf{(2)} We build a unified probing framework that reads out three complementary axes of spatial intelligence: semantic tagging, instance grouping, and 3D geometry prediction.
\textbf{(3)} We show a consistent division of strengths: VLMs make semantic categories and object instances more accessible, whereas VGMs make dense geometry and camera motion more accessible; simple feature fusion further indicates that these strengths are complementary rather than mutually exclusive.

\section{Related Work}

\paragraph{Vision-Language and Video Generation Backbones for Embodied AI}

VLMs have become common backbones for embodied agents, robot policies, and spatial reasoning systems, as language supervision encourages alignment with object categories, attributes, and instructions~\cite{liu2023visual,bai2025qwen2.5-VL,bai2025qwen3-VL,zhu2025internvl3,sensenova-si,yang2025thinking}.
In parallel, VGMs learn from temporally evolving visual observations and may acquire priors about dynamics and geometric consistency~\cite{wan2025wan,yang2025cogvideox,zheng2025open-sora,zhu2025aether,kong2024hunyuanvideo,genmo2024mochi}.

Recent embodied AI work has begun to compare or unify these two routes.
StarVLA supports VLM and VGM backbones under a shared VLA interface~\cite{community2026starvla}, and World Action Model studies compare video-based agents with VLA policies on manipulation, generalization, and robustness~\cite{ye2026worldactionmodelszeroshot,team2026motubrain,zhang2026world}.
Unlike these system-level evaluations, whose conclusions are entangled with many factors, we isolate the frozen perception substrate and compare what spatial information is already recoverable before downstream fine-tuning.

\paragraph{Frozen Probing for Spatial Perception}

Frozen-feature probing studies what information is already accessible in a representation by training only lightweight readouts~\cite{lewis2024does,cao2020behind}.
Our geometry probe is closely related to VidFM3D~\cite{huang2025vidfm3d}, which shows that frozen video foundation model features can support 3D geometry prediction; depth estimation with generative priors provides related evidence that VGMs features encode useful geometry~\cite{zhang2026dvd}.
However, these studies focus mainly on geometry and do not compare VLM and VGM pre-training schemes across semantic and object-centric axes. The key difference is that we place semantic tagging, instance grouping, and 3D geometry prediction under the probing framework, enabling a controlled comparison of VLMs and VGMs across three axes that is crucial for spatial intelligence.

\section{Probing Framework}

\begin{figure*}[t]
    \centering
    \includegraphics[width=\textwidth]{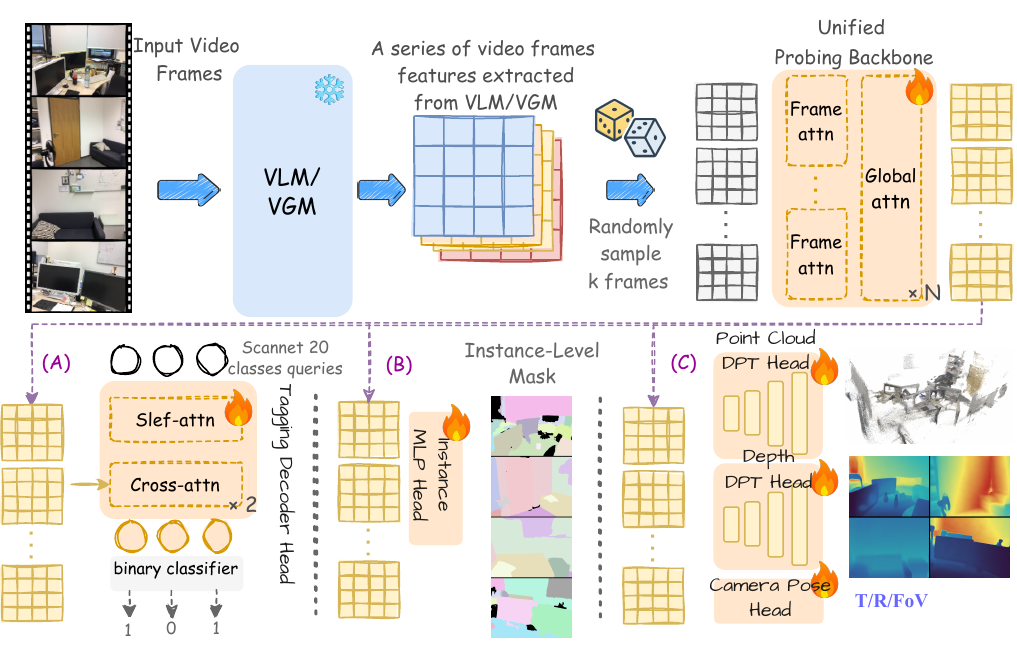}
    \caption{
        \textbf{Overview of our probing framework, where (A) is semantic tagging, (B) is instance grouping, and (C) is geometry prediction.}
        We freeze each VLM or VGM, extract temporally aligned video features, and train lightweight probes with an identical backbone architecture and task-specific heads.
        It is notable that the probing backbone is unified in architecture, but the three task probes are trained separately.
    }
    \label{fig:vlm-vgm-probe}
    \vspace{-3mm}
\end{figure*}

We study spatial intelligence through the lens of frozen representation probing.
Here, we define spatial intelligence as the extent to which semantic categories, object instances, and 3D scene geometry can be recovered from frozen visual representations.
Rather than fine-tuning foundation models, we freeze each Vision-Language Model (VLM) and Video Generation Model (VGM) and train unified lightweight readout modules on top of their intermediate video features.
As shown in Figure~\ref{fig:vlm-vgm-probe}, our framework first extracts temporally aligned frame features from each frozen model, then feeds a sampled set of frame features into a shared probing backbone, and finally attaches task-specific readout heads for semantic tagging, instance grouping, and 3D geometry prediction.
This setup allows us to focus on what information is already present in the frozen representations, rather than what can be added by task-specific fine-tuning or model-specific output designs.

\subsection{Frozen Feature Extraction}
Given a video \(V=(X_0,\ldots,X_{T-1})\), where \(X_i\) denotes an image frame, we randomly sample a start time \(t\in\{0,\ldots,T-76\}\) and take a contiguous 76-frame context
\(\mathcal{C}_t=(X_t,X_{t+1},\ldots,X_{t+75})\).
We use a fixed set of relative query indices
\(\mathcal{I}=\{0,1,5,9,\ldots,73\}\),
which gives a 20-frame query set
\(\mathcal{Q}_t=(X_t,X_{t+1},X_{t+5},X_{t+9},\ldots,X_{t+73})\).
That is, we keep the first two frames and then sample one frame every four frames within the context.
For each frozen model \(M\), we construct an aligned 20-position feature bank
\(\widetilde{\mathbf{F}}^{M}_{\ell}\in\mathbb{R}^{20\times H_M\times W_M\times C_M}\),
where \(H_M\times W_M\) and \(C_M\) denote the model-specific spatial grid and channel dimensions.
For VGMs, we feed the full context \(\mathcal{C}_t\) to the generator and use hidden activations from selected transformer blocks during a single denoising pass with a fixed timestep and empty text condition; image-to-video variants are also conditioned on the first frame.
The probe uses internal generator features rather than generated pixels.
Most video generators first encode the input clip with a spatiotemporal VAE~\cite{kingma2013auto} before the denoising transformer.
In WAN-style~\cite{wan2025wan} VAEs, the temporal encoder keeps the first frame as a separate latent slice and then compresses the remaining frames with a stride of four.
Thus a 76-frame context is represented by \(1+\lceil(76-1)/4\rceil=20\) latent slices, matching the temporal length of our query-frame convention.
We therefore take the hidden activation at these 20 latent positions as the VGM feature bank.
We refer readers to Appendix~\ref{app:wan-temporal-compression} for an implementation-level example of this temporal compression.
For VLMs, we directly feed the 20 query frames \(\mathcal{Q}_t\) to the model and collect visual-token hidden states from selected language-model layers.
These visual features are reshaped into frame-level grids, giving the same 20-frame feature-bank format.
The probing backbone then samples \(k\) frames from this bank as input.

\subsection{Unified Probing Backbone}
All tasks use lightweight probes with an identical backbone architecture.
Given the \(k\) sampled frame features
\(\mathbf{F}\in\mathbb{R}^{B\times k\times C_M\times H_f\times W_f}\),
we flatten each feature grid into \(P=H_f\times W_f\) patch tokens, project the model-specific channel dimension \(C_M\) to a shared width \(D\), and prepend one camera token to each frame.

The backbone is an \(N\)-layer alternating-attention transformer.
Each layer first performs \emph{frame attention}, where tokens interact within each frame, and then \emph{global attention}, where tokens interact across all sampled frames.
We concatenate the frame- and global-attention outputs at each layer, producing stage features
\(\mathbf{A}_n\in\mathbb{R}^{B\times k\times(P+1)\times2D}\).
For the 3D task, the point and depth heads use selected backbone stages for dense prediction, while the camera head uses the final-stage camera token.
For semantic tagging and instance grouping, we use only the final-stage patch tokens after removing the camera tokens.

\subsection{Task-Specific Readout Heads}
On top of this common backbone architecture, we attach task-specific heads for the three axes. The three tasks are trained independently.
Due to page limit, we describe only the core implementation here; more details are provided in Appendix~\ref{app:training-objectives}.

\paragraph{Semantic Tagging}
Semantic tagging asks whether each object category appears in the sampled frames, regardless of its location or extent.
Given the final backbone patch features
\(\mathbf{H}\in\mathbb{R}^{B\times k\times P\times 2D}\), the semantic head predicts a video-level multi-label vector over \(C\) object categories.
For a vocabulary \(\{c_1,\ldots,c_C\}\), we create one query per class, initialized from the CLIP text embedding of ``a photo of \(\{c_i\}\)'' when available.
These class queries serve as \(\mathbf{Q}\), while the flattened \(kP\) patch tokens from \(\mathbf{H}\) serve as \(\mathbf{K}\) and \(\mathbf{V}\), in a two-layer decoder with query self-attention followed by cross-attention to video tokens.
A per-class linear classifier maps each decoded query to one logit, yielding \(\mathbf{z}\in\mathbb{R}^{B\times C}\).
The binary target \(\mathbf{y}\in\{0,1\}^{B\times C}\) is computed from semantic masks over the sampled frames, so it reflects what the probe actually observes.
We train this head with asymmetric multi-label loss~\cite{ridnik2021asymmetric}, using negative focusing to suppress abundant easy negatives while preserving rare positive labels.

\paragraph{Instance Grouping}
Instance grouping asks which pixels across the sampled frames belong to the same object instance.
Given the final patch features \(\mathbf{H}\), we reshape them back to the feature grid and apply a small pixel-wise projection head followed by L2 normalization, producing instance embeddings
\(\mathbf{E}\in\mathbb{R}^{B\times k\times d_{\mathrm{ins}}\times H_f\times W_f}\).
We train \(\mathbf{E}\) with a multi-view contrastive pull-push loss~\cite{li2025iggt}: embeddings with the same instance ID are pulled together, while embeddings from different instances are pushed apart by a margin.
The loss is computed on sampled valid pixels rather than all \(kH_fW_f\) pixels, which avoids the quadratic cost of dense pairwise distances while preserving multi-view instance supervision.
At inference time, we cluster the normalized embeddings with HDBSCAN~\cite{mcinnes2017hdbscan} to obtain instance groups.

\paragraph{3D Geometry Prediction}
3D geometry prediction asks the probe to recover dense scene geometry and camera motion from the sampled features.
We use two DPT-style dense heads on selected backbone stages: one predicts a point map
\(\widehat{\mathbf{P}}\in\mathbb{R}^{B\times k\times H\times W\times3}\), and the other predicts a depth map
\(\widehat{\mathbf{D}}\in\mathbb{R}^{B\times k\times H\times W\times1}\); both heads also output confidence maps.
A camera head reads the final-stage camera token and predicts a pose encoding for each sampled frame.
The supervision comes from VGGT~\cite{wang2025vggt} generated point maps, depth maps, and camera poses, converted to a coordinate system where the first sampled frame is the reference.
We train point and depth prediction with confidence-weighted regression losses, and train camera prediction with a pose regression loss, which are similar with VGGT~\cite{wang2025vggt}.

\section{Experiments}

\begin{table*}[t]
\centering
\small
\setlength{\tabcolsep}{6pt}
\begin{tabular}{@{}l|ccc|cc|ccc@{}}
\toprule
& \multicolumn{3}{c|}{\textbf{Semantic Tagging}} & \multicolumn{2}{c|}{\textbf{Instance Grouping}} & \multicolumn{3}{c}{\textbf{3D Geometry}} \\
\cmidrule(lr){2-4}\cmidrule(lr){5-6}\cmidrule(lr){7-9}
\textbf{Model} & AP\(_{mid}\) \(\uparrow\) & mAP \(\uparrow\) & Mid Ratio \(\uparrow\) & T-SR \(\uparrow\) & T-mIoU \(\uparrow\) & P-map Err. \(\downarrow\) & AbsRel \(\downarrow\) & AUC@30 \(\uparrow\) \\
\midrule
\rowcolor{gray!15}\multicolumn{9}{c}{\textit{Video-Generation Models}} \\
WAN2.1-T2V-1.3B & 59.60 & 68.04 & 0.876 & 6.35 & 16.37 & 0.148 & 0.065 & 0.501 \\
WAN2.1-T2V-14B & 71.04 & 80.00 & 0.888 & 8.50 & 18.98 & \textbf{0.119} & \textbf{0.044} & \textbf{0.614} \\
WAN2.1-I2V-14B & 60.02 & 69.51 & 0.863 & 7.31 & 16.83 & 0.123 & 0.047 & 0.605 \\
CogVideoX-T2V-2B & 58.20 & 71.45 & 0.815 & 1.97 & 9.87 & 0.165 & 0.089 & 0.492 \\
CogVideoX-T2V-5B & 55.14 & 69.29 & 0.796 & 2.22 & 10.69 & 0.179 & 0.090 & 0.475 \\
CogVideoX-I2V-5B & 51.65 & 67.39 & 0.766 & 1.76 & 9.58 & 0.178 & 0.091 & 0.470 \\
OpenSora2.0 & 56.38 & 67.02 & 0.841 & 4.30 & 13.07 & 0.141 & 0.066 & 0.560 \\
Aether & 57.02 & 66.43 & 0.858 & 2.39 & 10.54 & 0.165 & 0.084 & 0.495 \\
\rowcolor{green!8}\textit{Avg.} & 58.63 & 69.89 & 0.838 & 4.35 & 13.24 & 0.152 & 0.072 & 0.527 \\
\midrule
\rowcolor{gray!15}\multicolumn{9}{c}{\textit{Vision-Language Models}} \\
InternVL3-1B & 87.53 & 92.52 & 0.946 & 10.45 & 20.79 & 0.262 & 0.125 & 0.254 \\
InternVL3-2B & 86.88 & 92.34 & 0.941 & 10.02 & 21.25 & 0.221 & 0.118 & 0.323 \\
InternVL3-8B & 89.00 & 93.17 & 0.955 & 11.22 & 22.60 & 0.202 & 0.106 & 0.362 \\
InternVL3.5-4B & 87.98 & 91.84 & 0.958 & 10.18 & 21.74 & 0.214 & 0.113 & 0.338 \\
InternVL3.5-8B & 87.49 & 91.65 & 0.955 & 10.83 & 22.02 & 0.205 & 0.110 & 0.361 \\
Qwen2.5-VL-3B & 84.04 & 90.59 & 0.928 & 10.12 & 21.53 & 0.320 & 0.152 & 0.163 \\
Qwen2.5-VL-7B & 87.64 & 91.62 & 0.957 & 10.79 & 22.21 & 0.234 & 0.127 & 0.310 \\
Qwen3-VL-2B & \textbf{90.14} & \textbf{93.56} & \textbf{0.963} & \textbf{13.56} & \textbf{25.50} & 0.202 & 0.096 & 0.371 \\
Qwen3-VL-4B & 88.62 & 92.47 & 0.958 & 12.70 & 24.62 & 0.191 & 0.093 & 0.390 \\
Qwen3-VL-8B & 83.49 & 91.08 & 0.917 & 12.41 & 24.27 & 0.180 & 0.091 & 0.424 \\
\rowcolor{green!8}\textit{Avg.} & 87.28 & 92.08 & 0.948 & 11.23 & 22.66 & 0.223 & 0.113 & 0.330 \\
\bottomrule
\end{tabular}
\caption{
Main comparison of frozen representations under the unified probe.
Semantic and instance metrics are reported as percentages; 3D metrics are reported on their original scales.
Mid Ratio is dimensionless.
Average rows report the unweighted mean within each model family.
Best results are shown in bold.
}
\label{tab:main-results}
\vspace{-3mm}
\end{table*}

\subsection{Experimental Setup}

\paragraph{Dataset for training and evaluation.}
We evaluate the three probing axes on two datasets.
For semantic tagging, we use ScanNet20~\cite{dai2017scannet} and predict which object categories appear in the sampled frames.
For instance grouping, we use ScanNet multi-view instance masks and evaluate whether pixels belonging to the same object are grouped consistently across views.
For both ScanNet tasks, we use the official training split for probe training and the official validation split for evaluation.
For 3D geometry, we use DL3DV~\cite{ling2024dl3dv} and supervise the probe with VGGT~\cite{wang2025vggt}-generated point maps, depth maps, and camera poses.
Following ~\cite{huang2025vidfm3d}, we use the first 6K DL3DV samples and split them into training and test sets with a 9:1 ratio.

\paragraph{Models.}
We compare two families of frozen foundation models.
The VGM side includes WAN~\cite{wan2025wan} variants, CogVideoX~\cite{yang2025cogvideox} variants, OpenSora-2.0~\cite{zheng2025open-sora}, and Aether~\cite{zhu2025aether}.
The VLM side includes InternVL3~\cite{zhu2025internvl3}, InternVL3.5~\cite{wang2025internvl3-5}, Qwen2.5-VL~\cite{bai2025qwen2.5-VL}, and Qwen3-VL~\cite{bai2025qwen3-VL} variants.

\paragraph{Protocol.}
All experiments use the same 76-frame context and the same temporally aligned feature-bank construction described in the previous section.
The geometry task samples \(k=4\) frames, while semantic tagging and instance grouping sample \(k=8\) frames.
We use fixed feature layers for each model family and task.
We train geometry, instance grouping, and semantic tagging probes for 60, 40, and 10 epochs, respectively.

\paragraph{Metrics.}
For semantic tagging, we report mAP, the macro average precision over ScanNet20 categories, $\text{AP}_{mid}$, the average precision on less frequent object categories, and Mid Ratio, defined as $\text{AP}_{mid}/\text{mAP}$, which measures whether performance extends beyond common categories.
For instance grouping, we report T-mIoU, the mean IoU between each ground-truth instance and its best predicted cluster over aggregated views, and T-SR, the fraction of ground-truth instances successfully grouped in every view where they appear.
For 3D geometry, we report P-map Err., depth AbsRel, and camera AUC@30.
P-map Err. denotes aligned point-map error, measuring the mean error after similarity alignment to VGGT point maps; depth AbsRel measures relative depth error; and AUC@30 measures relative camera-pose accuracy up to a 30-degree threshold.

\paragraph{More details} are described in Appendix~\ref{app:implementation-details}.

\subsection{Main Results}

\paragraph{Semantic Tagging}
As shown in Table~\ref{tab:main-results}, VLMs clearly outperform VGMs on video-level object recognition.
On average, VLMs improve mAP from 69.89 to 92.08 and AP\(_{mid}\) from 58.63 to 87.28.
This gap suggests that language-aligned visual representations preserve substantially more object-category information than representations learned only from video generation objectives.
The difference is even clearer in Mid Ratio: VLMs reach an average ratio of 0.948, compared with 0.838 for VGMs.
Thus, the VLM advantage is not only due to recognizing common ScanNet categories; their AP on less frequent categories stays much closer to their overall mAP.

Among VGMs, WAN2.1-T2V-14B is the strongest semantic tagger, indicating that scale helps, but it still remains below all evaluated VLMs.
Among VLMs, Qwen3-VL-2B obtains the best AP\(_{mid}\), mAP, and Mid Ratio, but most VLMs cluster in a narrow high-performing range, showing that the semantic advantage is a family-level trend rather than a single-checkpoint outlier.

\vspace{-3mm}
\paragraph{Instance Grouping}
Instance grouping shows a similar, though less saturated, trend.
VLMs improve the family average from 13.24 to 22.66 in T-mIoU and from 4.35 to 11.23 in T-SR.
Although grouping pixels across views requires spatial consistency, these results indicate that the task also strongly benefits from object-centric semantics: features must identify which regions form the same object, not merely which pixels are geometrically nearby.
This suggests that instance grouping is not a pure geometry probe, but also reflects the quality of object-centric representations.

The strongest model is again Qwen3-VL-2B, with 25.50 T-mIoU and 13.56 T-SR, while Qwen3-VL-4B and Qwen3-VL-8B remain close behind.
Among VGMs, WAN2.1-T2V-14B performs best, suggesting that stronger video generation models do encode useful cross-view grouping signals.
However, even the best VGM remains below the VLM average, showing that video-generation pre-training alone does not provide the same level of object-level separability as vision-language alignment.

\begin{table}[t]
\centering
\small
\setlength{\tabcolsep}{1.5pt}
\begin{tabular}{@{}lcccc@{}}
\toprule
\textbf{Model} & mAP \(\uparrow\) & T-mIoU \(\uparrow\) & AbsRel \(\downarrow\) & AUC@30 \(\uparrow\) \\
\midrule
WAN2.1-T2V-14B & 80.00 & 18.98 & 0.044 & 0.614 \\
Qwen3-VL-8B & 91.08 & 24.27 & 0.091 & 0.424 \\
Fusion & \textbf{92.30} & \underline{23.70} & \textbf{0.042} & \textbf{0.615} \\
\bottomrule
\end{tabular}
\caption{
Comparison among WAN2.1-T2V-14B, Qwen3-VL-8B, and their feature-level fusion version.
The fusion model concatenates normalized frozen features before the same probe backbone. \textbf{Bold} numbers indicate the best performance, while \underline{underlined} numbers indicate the second best.
}
\label{tab:fusion-results}
\end{table}

\paragraph{3D Geometry}
The 3D Geometry task reverses the trend observed above.
VGMs achieve better family averages on all three geometry metrics, with lower P-map Err. (0.152 vs. 0.223), lower AbsRel (0.072 vs. 0.113), and higher AUC@30 (0.527 vs. 0.330).
This suggests that video-generation representations make dense geometry and camera information more directly recoverable, likely because generation requires maintaining spatial consistency across frames.
This observation is consistent with recent studies showing that video generation features can support direct 3D prediction and depth estimation through their implicit geometric priors~\cite{huang2025vidfm3d,zhang2026dvd}.
WAN2.1-T2V-14B obtains the best geometry scores among all models, with WAN2.1-I2V-14B close behind.
This indicates that the WAN representations contain particularly accessible multi-view geometry, beyond their weaker semantic and instance-grouping results.

Among VLMs, Qwen3-VL-8B is the strongest geometry model, reaching 0.180 P-map Err. and 0.424 AUC@30.
This may be related to Qwen3-VL's explicit spatial and 3D training, as its technical report~\cite{bai2025qwen3-VL} describes.
Still, it remains below the VGM average, suggesting that such 3D-aware supervision helps but does not fully replace the geometric bias induced by video-generation pre-training.

\paragraph{Takeaways}
Overall, Table~\ref{tab:main-results} shows a clear division of strengths between the two model families.
VLMs provide stronger semantic and object-centric representations, whereas VGMs provide more accessible dense geometric signals.
Thus, spatial intelligence in current foundation models is not captured by a single axis: language alignment favors object semantics, while video generation favors multi-frame geometric structure.
This complementarity motivates the evaluation of spatial intelligence with multiple probing tasks rather than relying only on recognition or only on reconstruction.

\subsection{Naive Feature Fusion can Bridge Semantic and Geometric Strengths}
\label{sec:fusion}

\begin{table}[t]
\centering
\small
\setlength{\tabcolsep}{3pt}
\begin{tabular}{@{}llcl@{}}
\toprule
\textbf{Task} & \textbf{Metric} & \textbf{Depth} & \textbf{Ranking} \\
\midrule
Instance & T-mIoU \(\uparrow\) & 1 & Q \(>\) I \(>\) W \(>\) C \\
Instance & T-mIoU \(\uparrow\) & 2 & Q \(>\) I \(>\) W \(>\) C \\
Instance & T-mIoU \(\uparrow\) & 4 & Q \(>\) I \(>\) W \(>\) C \\
Instance & T-mIoU \(\uparrow\) & 6 & Q \(>\) I \(>\) W \(>\) C \\
\midrule
3D & P-map Err. \(\downarrow\) & 1 & W \(>\) C \(>\) Q \(>\) I \\
3D & P-map Err. \(\downarrow\) & 2 & W \(>\) C \(>\) Q \(>\) I \\
3D & P-map Err. \(\downarrow\) & 4 & W \(>\) C \(>\) Q \(>\) I \\
3D & P-map Err. \(\downarrow\) & 6 & W \(>\) C \(>\) Q \(>\) I \\
\bottomrule
\end{tabular}
\caption{
Probe-depth ablation on representative models.
W, C, I, and Q denote WAN2.1-T2V-14B, CogVideoX-I2V-5B, InternVL3-8B, and Qwen3-VL-8B, respectively.
For P-map Err., lower values are ranked higher.
Full numerical results are provided in Appendix~\ref{app:probe-depth-ablation}.
}
\label{tab:probe-depth-ablation}
\vspace{-3mm}
\end{table}

The complementarity in Table~\ref{tab:main-results} raises a natural follow-up question:
can VLM and VGM strengths be combined in a single representation?
As a first test, we evaluate an intentionally naive feature-level fusion baseline: interpolate WAN2.1-T2V-14B features to the temporal length of Qwen3-VL-8B, normalize each model's features independently, concatenate them along channels, and feed the fused features into the same probe.

Table~\ref{tab:fusion-results} shows that even this simple fusion produces a representation strong in both semantics and geometry.
Relative to WAN2.1-T2V-14B, it substantially improves semantic tagging mAP and instance grouping, approaching Qwen3-VL-8B on grouping and surpassing it on tagging.
Relative to Qwen3-VL-8B, it lowers depth AbsRel and raises camera AUC@30, slightly exceeding WAN2.1-T2V-14B on both geometry metrics.
Thus, feature fusion can preserve VLM semantic strength while matching or improving VGM geometry.

These results further support that VLMs and VGMs provide complementary spatial information.
The clear gain from such a simple fusion suggests that stronger fusion mechanisms may be an especially promising direction.
Future work can explore principled ways to integrate VLM object-level semantics with the dense geometric structure learned by VGMs, toward stronger spatial intelligence.

\begin{figure}[t]
    \centering
    \includegraphics[width=\columnwidth]{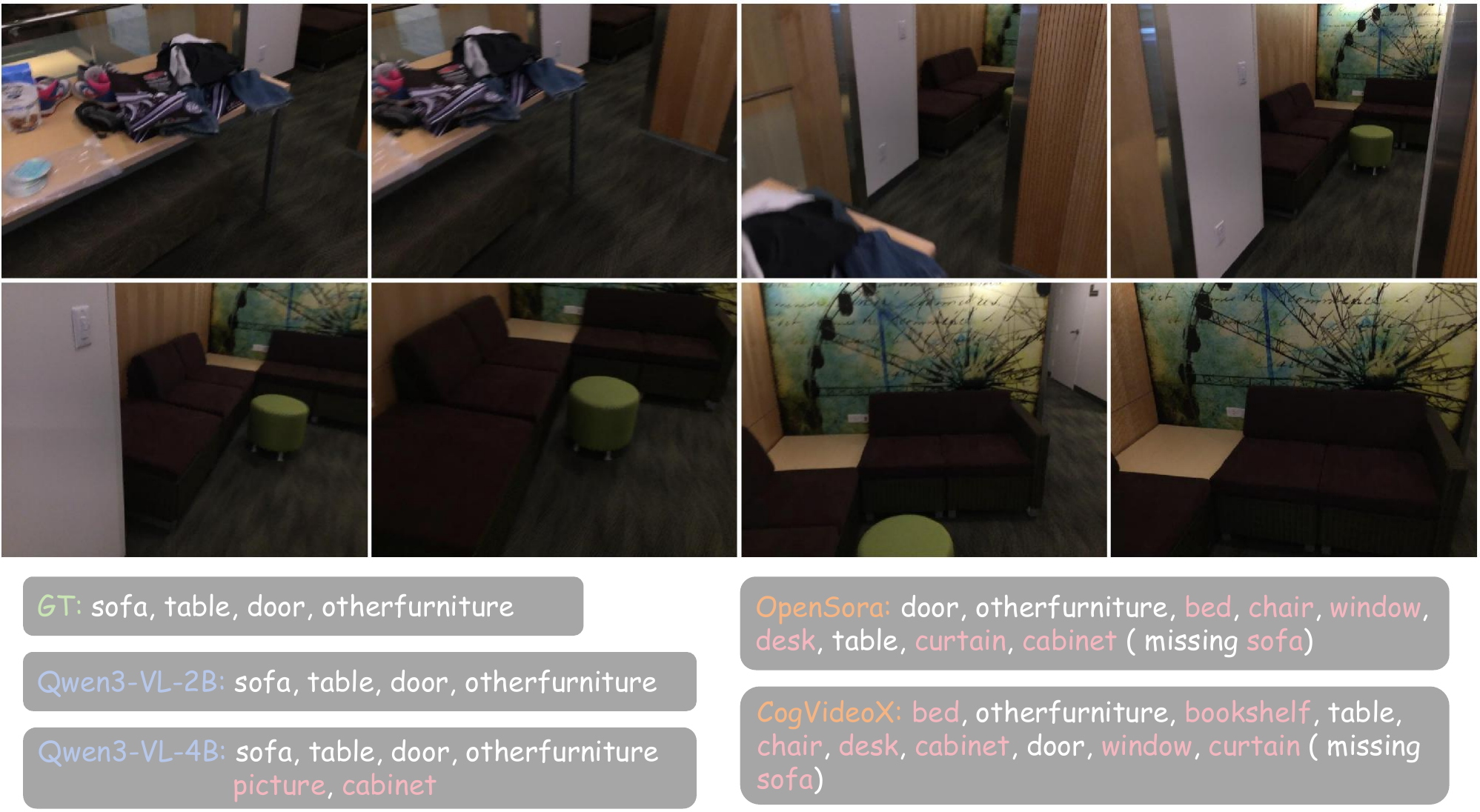}
    \caption{
        Qualitative semantic tagging on ScanNet scene0559\_01.
    }
    \label{fig:case-tagging-scene0559}
\end{figure}

\begin{figure}[t]
    \centering
    \includegraphics[width=\columnwidth]{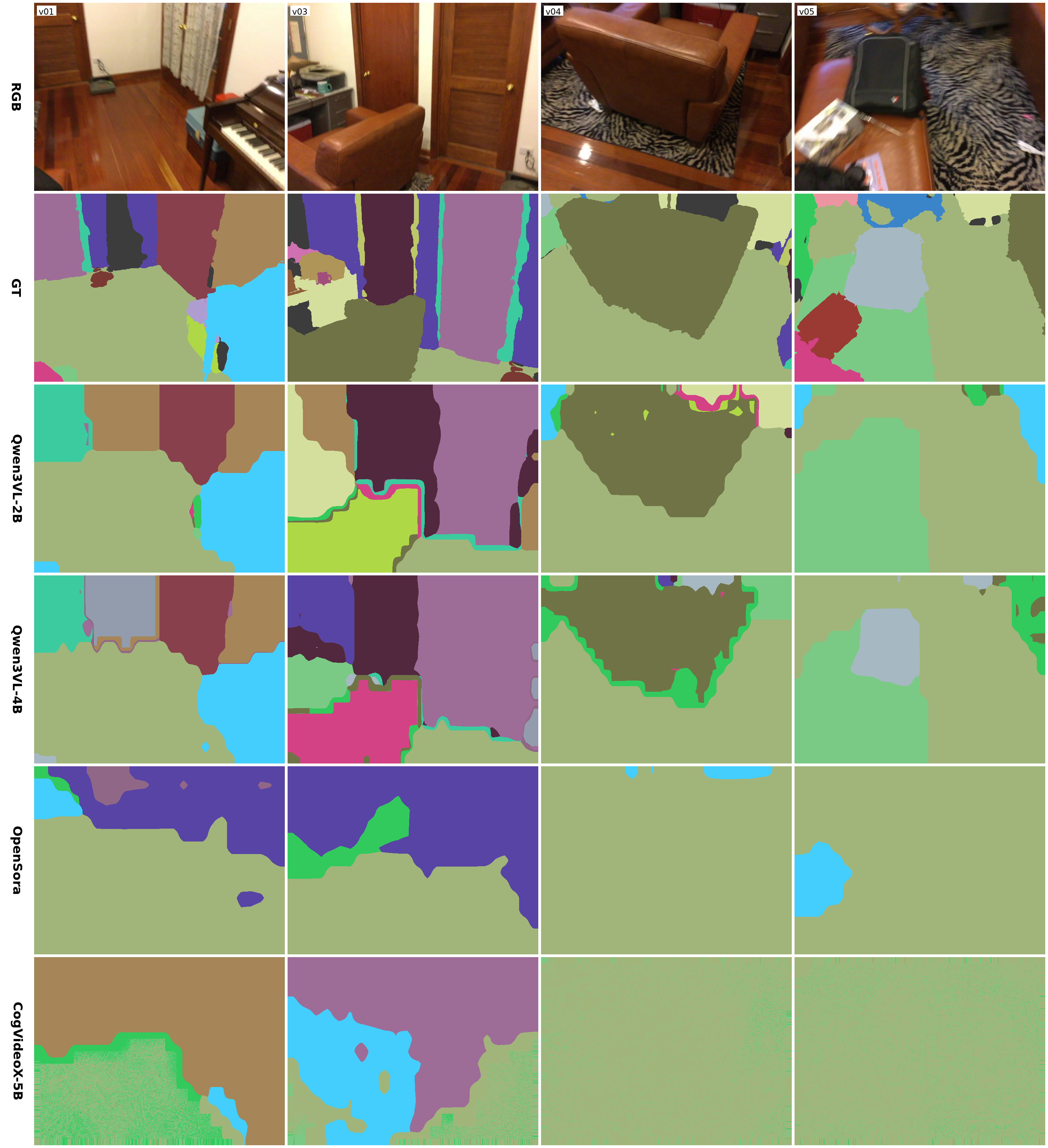}
    \caption{
        Qualitative instance grouping on ScanNet scene0050\_00.
    }
    \label{fig:case-instance-scene0050}
    \vspace{-4mm}
\end{figure}

\subsection{Ablation Study}

Table~\ref{tab:probe-depth-ablation} varies the depth of the lightweight probe while keeping the frozen features fixed.
Across depths, the relative ranking of representative models remains stable for both instance grouping and 3D geometry.
This suggests that the probe is not simply learning these tasks from scratch; rather, it mainly reads out semantic, instance-level, and geometric information already encoded in the frozen representations.

\subsection{Case Study}

\begin{figure}[t]
    \centering
    \includegraphics[width=\columnwidth]{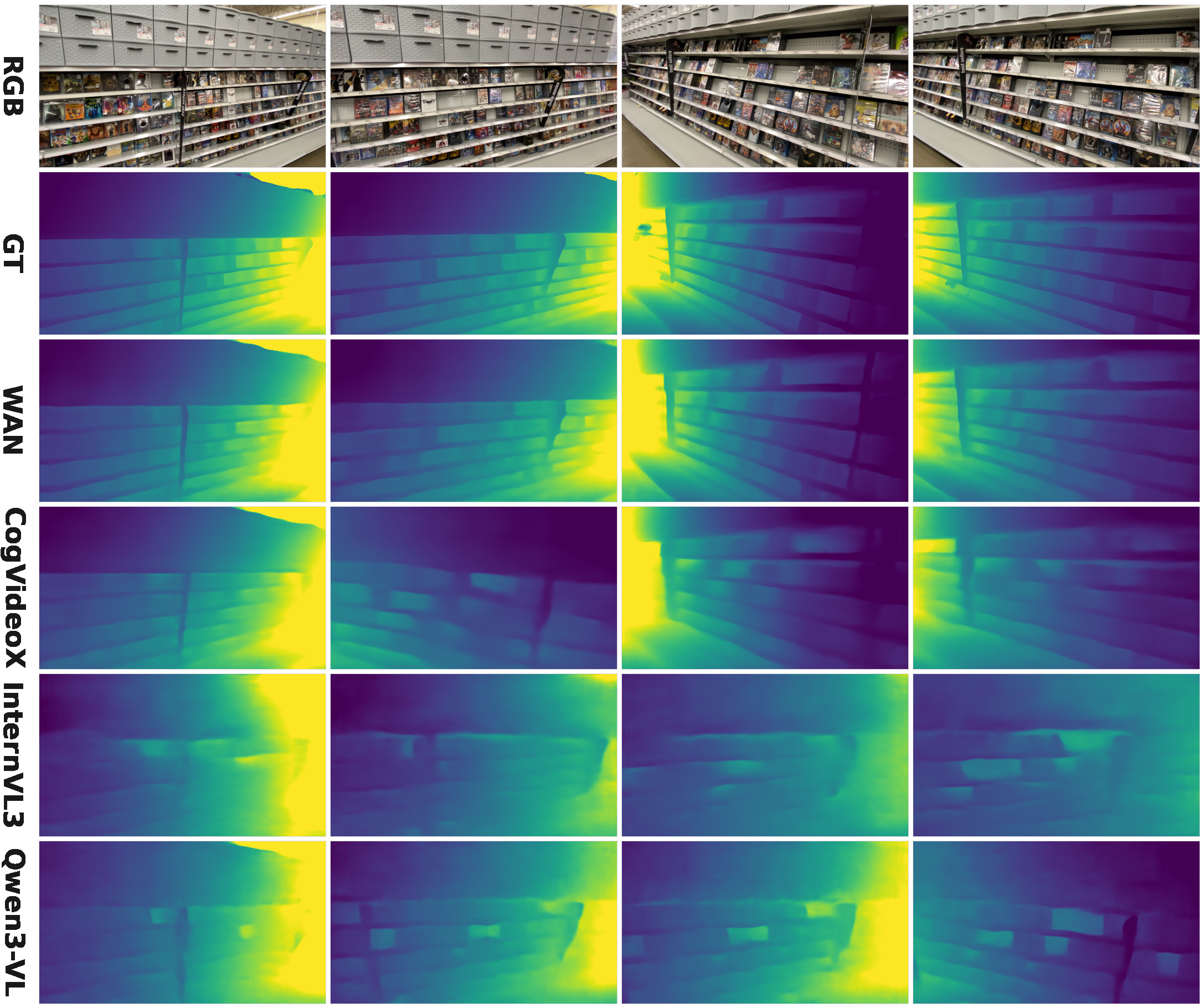}
    \caption{
        Qualitative depth prediction on a DL3DV case.
    }
    \label{fig:case-depth-batch0000}
    \vspace{-4mm}
\end{figure}

\paragraph{Semantic Tagging}  As shown in Figure~\ref{fig:case-tagging-scene0559}, both Qwen3-VL models recall all positive labels, while all VGM probes miss the sofa, a key object in the scene, revealing their weaker semantic recognition ability.

\paragraph{Instance Grouping} As shown in Figure~\ref{fig:case-instance-scene0050}, VLM features separate major objects such as the sofa and door more clearly, while VGM features often merge them into coarse regions, suggesting that VGM features are less able to distinguish different semantic entities.
An additional example is provided in Appendix Figure~\ref{fig:case-instance-scene0030}.

\paragraph{3D Geometry} As shown in Figure~\ref{fig:case-depth-batch0000}, VGM features recover sharper depth structure and preserve the shelf geometry more faithfully than VLM features.
The VLM predictions capture the coarse scene layout, blur object boundaries and local depth changes. Another point cloud example (in Appendix Figure~\ref{fig:case-pointcloud-batch0009}) also illustrates the observation that VLMs have weak 3D geometric ability. Compared with the reconstructed point clouds from VGMs, the VLM point clouds are much noisier and lack clear structure.
Additional depth and point-cloud examples are provided in Appendix Figures~\ref{fig:case-depth-batch0005}.

\section{Conclusion}

We present a unified frozen-feature probing framework for comparing VLM and VGM representations across semantic tagging, instance grouping, and 3D geometry prediction. Our results show that spatial intelligence is not captured by a single axis: VLMs provide stronger semantic and object-centric representations, while VGMs make dense geometry and camera motion more recoverable. Probe-depth ablations show that these trends are stable across readout capacities, suggesting that the probes mainly expose information already encoded in frozen features. Finally, simple feature-level fusion combines the strengths of both families, improving semantic performance while preserving strong geometry. These findings suggest that future spatial-understanding backbones may benefit from integrating language-aligned object semantics with video-generation geometric priors.

\section{Limitations}

Our study has several limitations. First, although semantic tagging, instance grouping, and 3D geometry cover important axes of spatial intelligence, they do not exhaust the full space of spatial reasoning; capabilities such as physical dynamics, affordance understanding, active exploration, and long-horizon embodied reasoning are left for future work. Second, our evaluation is conducted mainly on ScanNet and DL3DV, which emphasize indoor scenes and reconstructed video data, so the conclusions may not fully transfer to outdoor, highly dynamic, or robot-collected environments. Third, the comparison necessarily depends on practical design choices, including selected feature layers, frame sampling, spatial resolution, and VGM denoising timesteps. We mitigate this concern with controlled protocols and probe-depth ablations, but a broader sensitivity analysis would further strengthen the conclusions. Finally, our feature-level fusion experiment is intentionally simple and should be viewed as a proof of concept rather than a final fusion architecture; more principled mechanisms for aligning and integrating VLM and VGM representations remain an important direction for future work.


\bibliography{custom}

\appendix

\section*{Appendix}
\section{VGM Temporal Compression}
\label{app:wan-temporal-compression}

We use WAN as a representative example to explain how a video generator converts several input frames into one latent temporal position.
WAN first encodes the RGB video with a causal spatiotemporal VAE before the denoising transformer.
The VAE treats the first frame as a separate causal slice, and the remaining frames are processed in temporal chunks.
Inside the encoder, temporal compression is implemented by two 3D downsampling blocks.
Each block contains a causal temporal convolution with kernel size \((3,1,1)\) and stride \((2,1,1)\), so the two blocks together give an effective temporal stride of \(2\times2=4\).
Therefore, after the first latent slice, each latent temporal step advances by four input frames.

Concretely, WAN feature extraction uses an 81-frame window.
The VAE produces
\begin{equation}
T_{\mathrm{lat}}
=
1+\frac{81-1}{4}
=21
\end{equation}
latent temporal positions: one position for the first frame and 20 positions for the remaining 80 frames.
Our probing context uses the first 76 frames of this window.
Accordingly, we retain the first
\begin{equation}
1+\left\lceil\frac{76-1}{4}\right\rceil
=20
\end{equation}
latent positions and discard the last WAN position, which would correspond to frames beyond the 76-frame context.
This gives a 20-position VGM feature bank aligned with the query-frame convention used by the VLM features.

\section{Detailed Training Objectives}
\label{app:training-objectives}

This section provides the training objectives used by the three probes.
All foundation models are frozen; only the shared probing backbone and the corresponding task head are optimized.
Given sampled frozen features, the probing backbone outputs stage tokens
\(\{\mathbf{A}_n\}_{n=1}^{N}\), where each stage has shape
\(\mathbf{A}_n\in\mathbb{R}^{B\times k\times(P+1)\times2D}\) after concatenating the frame- and global-attention outputs.
For tagging and instance grouping, we use the final-stage patch tokens after removing the camera token.
For geometry, the dense heads use selected stages, and the camera head uses the final-stage camera token.

\paragraph{Semantic tagging.}
For a sampled ScanNet video, let \(n_{t,c}\) denote the number of visible pixels of class \(c\) in sampled frame \(t\).
The binary target is constructed from the sampled frames rather than from the full clip:
\begin{equation}
y_c =
\mathbf{1}\left[
\sum_{t=1}^{k} n_{t,c} \ge \tau_{\mathrm{pix}}
\;\wedge\;
\sum_{t=1}^{k}\mathbf{1}[n_{t,c}>0] \ge \tau_{\mathrm{frm}}
\right],
\end{equation}
where \(\tau_{\mathrm{pix}}=200\) and \(\tau_{\mathrm{frm}}=1\) in our experiments.
This makes the label reflect what the probe actually observes.

The semantic head produces one logit \(z_c\) for each class, with probability
\(p_c=\sigma(z_c)\).
We optimize the asymmetric multi-label loss~\cite{ridnik2021asymmetric}.
Let \(\tilde{p}^{-}_c=\min(1,1-p_c+m)\), where \(m\) is the optional negative-probability shift; we use \(m=0\), i.e., no probability clipping.
Define
\begin{equation}
p^t_c = y_c p_c + (1-y_c)\tilde{p}^{-}_c,\quad
\gamma_c = y_c\gamma_{\mathrm{pos}} + (1-y_c)\gamma_{\mathrm{neg}}.
\end{equation}
The per-sample tagging loss is
\begin{equation}
\mathcal{L}_{\mathrm{tag}}
=
-\sum_{c=1}^{C}
(1-p^t_c)^{\gamma_c}
\left[
y_c\log p_c + (1-y_c)\log \tilde{p}^{-}_c
\right].
\end{equation}
The batch loss is the mean of this quantity over all samples.
We set \(\gamma_{\mathrm{neg}}=4\) and \(\gamma_{\mathrm{pos}}=0\), so easy negatives are down-weighted while rare positives are not focal-suppressed.

\paragraph{Instance grouping.}
The instance head maps the final patch tokens to per-pixel embeddings
\(\mathbf{e}_{t,u}\in\mathbb{R}^{d_{\mathrm{ins}}}\), followed by L2 normalization.
During training, embeddings are bilinearly upsampled to the ground-truth mask resolution and normalized again.
Let \(m_{t,u}\) be the ScanNet instance ID at pixel \(u\) in frame \(t\), and let \(\mathcal{V}\) be the set of valid pixels after removing ignored IDs such as background.
For each scene, we sample a set \(\Omega\subset\mathcal{V}\) of \(P_s=2048\) valid pixels.
For two sampled pixels \(i,j\in\Omega\), define
\begin{equation}
\begin{aligned}
d_{ij} &= \|\mathbf{e}_i-\mathbf{e}_j\|_2,\\
\mathcal{P} &= \{(i,j):m_i=m_j,\ i\neq j\},\\
\mathcal{N} &= \{(i,j):m_i\neq m_j\}.
\end{aligned}
\end{equation}
The multi-view contrastive pull-push loss is
\begin{equation}
\mathcal{L}_{\mathrm{pull}}
=
\frac{1}{|\mathcal{P}|}\sum_{(i,j)\in\mathcal{P}} d_{ij},
\end{equation}
\begin{equation}
\mathcal{L}_{\mathrm{push}}
=
\frac{1}{|\mathcal{N}|}\sum_{(i,j)\in\mathcal{N}}
\max(0,\mu-d_{ij}),
\end{equation}
and
\begin{equation}
\mathcal{L}_{\mathrm{ins}}
=
\lambda_{\mathrm{pull}}\mathcal{L}_{\mathrm{pull}}
+
\lambda_{\mathrm{push}}\mathcal{L}_{\mathrm{push}}.
\end{equation}
We use margin \(\mu=1.0\) and \(\lambda_{\mathrm{pull}}=\lambda_{\mathrm{push}}=1\).
At evaluation time, the normalized embeddings are clustered with HDBSCAN, and the resulting clusters are compared with the ground-truth instance masks.

\paragraph{3D geometry prediction.}
The geometry probe predicts a point map \(\widehat{\mathbf{P}}\), a depth map \(\widehat{\mathbf{D}}\), and a sequence of camera-pose encodings
\(\{\widehat{\mathbf{q}}^{(r)}\}_{r=1}^{R}\).
The supervision comes from VGGT-generated point maps, depth maps, confidence maps, intrinsics, and extrinsics.
Before training, poses, point maps, and depth maps are converted to the first sampled frame as the reference coordinate system, and the scene is scaled by the average reference-frame point distance.

For point-map and depth prediction, we use the same confidence-weighted regression form.
Let \(\widehat{\mathbf{Y}}\) and \(\mathbf{Y}\) denote either predicted and target point maps or predicted and target depth maps.
Let \(C_{t,u}\) be the VGGT confidence and \(M_{t,u}\) the optional valid foreground mask.
The regression term is
\begin{equation}
\mathcal{L}_{\mathrm{reg}}(\widehat{\mathbf{Y}},\mathbf{Y})
=
\frac{
\sum_{t,u} M_{t,u} C_{t,u}
\left\|\widehat{\mathbf{Y}}_{t,u}-\mathbf{Y}_{t,u}\right\|_2
}{
\sum_{t,u} M_{t,u} + \epsilon
}.
\end{equation}
For point maps, both prediction and target are additionally normalized by their valid-pixel average distance before computing this loss; depth maps use the scaled depths directly.
The implementation also computes a multi-scale gradient regularizer,
\begin{equation}
\begin{aligned}
\mathcal{L}_{\mathrm{grad}}
&=
\frac{1}{S_g}\sum_{s=0}^{S_g-1}
\left(
\|\nabla_x \Delta^{(s)}\|_1
+\|\nabla_y \Delta^{(s)}\|_1
\right),\\
\Delta&=\widehat{\mathbf{Y}}-\mathbf{Y}.
\end{aligned}
\end{equation}
where scale \(s\) subsamples the image grid by stride \(2^s\).
In the main experiments, the gradient-loss weights are set to zero, so this term is logged but not included in the optimized objective.

For camera prediction, the target pose encoding
\(\mathbf{q}\in\mathbb{R}^{9}\) is derived from the reference-frame extrinsics and intrinsics using the VGGT pose parameterization, containing translation, quaternion rotation, and field-of-view terms.
The camera head performs iterative refinement and outputs \(R=4\) predictions.
For iteration \(r\), we compute Huber losses on the three pose components:
\begin{equation}
\begin{aligned}
\mathcal{L}^{(r)}_{\mathrm{cam}}
&=
\ell_{\mathrm{Huber}}(\widehat{\mathbf{q}}^{(r)}_{T},\mathbf{q}_{T})
\\
&\quad+
\ell_{\mathrm{Huber}}(\widehat{\mathbf{q}}^{(r)}_{R},\mathbf{q}_{R})
+
\frac{1}{2}
\ell_{\mathrm{Huber}}(\widehat{\mathbf{q}}^{(r)}_{F},\mathbf{q}_{F}).
\end{aligned}
\end{equation}
Later refinement steps receive larger weights:
\begin{equation}
\mathcal{L}_{\mathrm{cam}}
=
\frac{1}{R}
\sum_{r=1}^{R}
\gamma^{R-r}\mathcal{L}^{(r)}_{\mathrm{cam}},
\quad \gamma=0.6.
\end{equation}
The final geometry training loss is
\begin{equation}
\mathcal{L}_{\mathrm{geo}}
=
\lambda_P\mathcal{L}_{\mathrm{reg}}(\widehat{\mathbf{P}},\mathbf{P})
+
\lambda_D\mathcal{L}_{\mathrm{reg}}(\widehat{\mathbf{D}},\mathbf{D})
+
\lambda_C\mathcal{L}_{\mathrm{cam}},
\end{equation}
with \(\lambda_P=\lambda_D=\lambda_C=1\) in all main geometry experiments.

\begin{table*}[t]
\centering
\small
\setlength{\tabcolsep}{4pt}
\begin{tabular}{@{}llcrrrrl@{}}
\toprule
\textbf{Task} & \textbf{Metric} & \textbf{Depth} & \textbf{WAN} & \textbf{Cog} & \textbf{Intern} & \textbf{Qwen} & \textbf{Ranking} \\
\midrule
Instance grouping & T-mIoU \(\uparrow\) & 1 & 19.29 & 9.94 & 22.83 & 24.61 & Qwen \(>\) Intern \(>\) WAN \(>\) Cog \\
Instance grouping & T-mIoU \(\uparrow\) & 2 & 18.98 & 9.58 & 22.60 & 24.27 & Qwen \(>\) Intern \(>\) WAN \(>\) Cog \\
Instance grouping & T-mIoU \(\uparrow\) & 4 & 18.36 & 9.43 & 21.55 & 23.36 & Qwen \(>\) Intern \(>\) WAN \(>\) Cog \\
Instance grouping & T-mIoU \(\uparrow\) & 6 & 18.82 & 8.81 & 21.56 & 23.97 & Qwen \(>\) Intern \(>\) WAN \(>\) Cog \\
\midrule
3D geometry & P-map Err. \(\downarrow\) & 1 & 0.126 & 0.187 & 0.214 & 0.202 & WAN \(>\) Cog \(>\) Qwen \(>\) Intern \\
3D geometry & P-map Err. \(\downarrow\) & 2 & 0.119 & 0.173 & 0.204 & 0.191 & WAN \(>\) Cog \(>\) Qwen \(>\) Intern \\
3D geometry & P-map Err. \(\downarrow\) & 4 & 0.119 & 0.178 & 0.202 & 0.180 & WAN \(>\) Cog \(>\) Qwen \(>\) Intern \\
3D geometry & P-map Err. \(\downarrow\) & 6 & 0.121 & 0.174 & 0.204 & 0.182 & WAN \(>\) Cog \(>\) Qwen \(>\) Intern \\
\bottomrule
\end{tabular}
\caption{
Full numerical results for the probe-depth ablation.
WAN, Cog, Intern, and Qwen denote WAN2.1-T2V-14B, CogVideoX-I2V-5B, InternVL3-8B, and Qwen3-VL-8B, respectively.
Instance grouping is reported with T-mIoU in percentages; 3D geometry is reported with P-map Err. on its original scale.
For P-map Err., lower values are ranked higher.
}
\label{tab:probe-depth-ablation-full}
\end{table*}

\begin{figure*}[t]
    \centering
    \includegraphics[width=\textwidth]{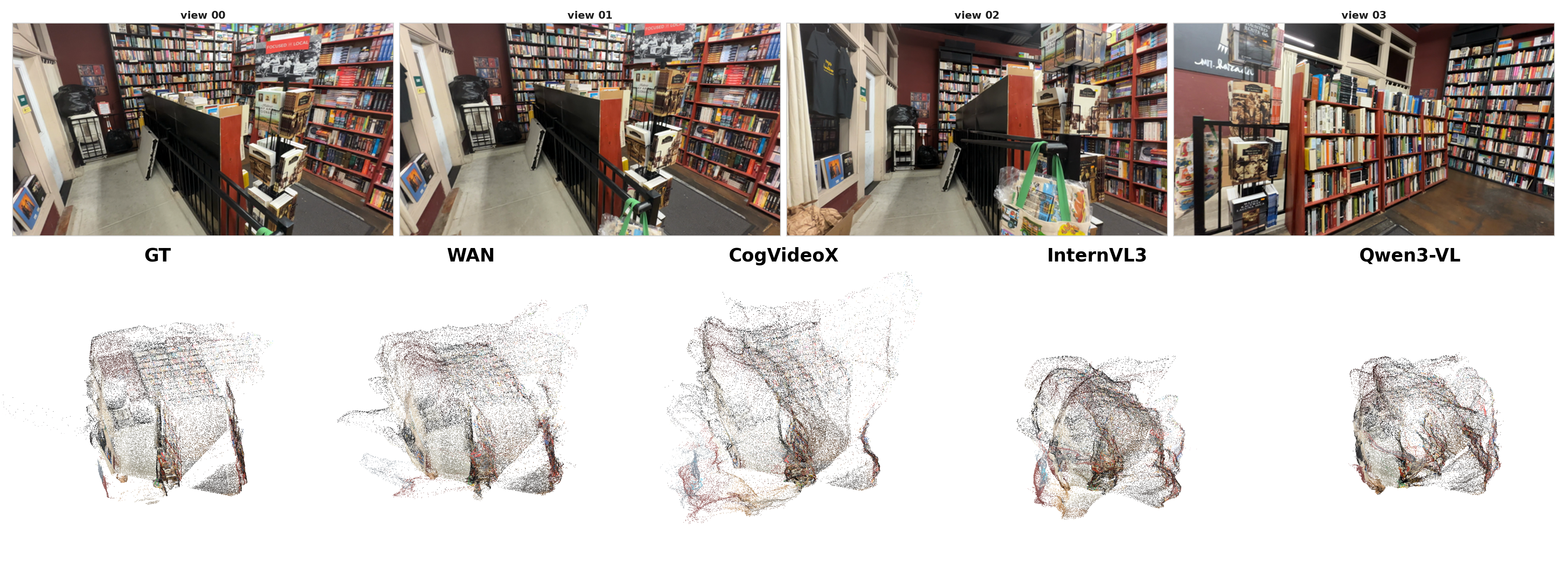}
    \caption{
        Point-cloud visualization on a DL3DV bookstore scene.
        The top row shows four input RGB views, and the bottom row compares GT, WAN, CogVideoX, InternVL3, and Qwen3-VL point clouds from a shared viewpoint.
    }
    \label{fig:case-pointcloud-batch0009}
\end{figure*}

\section{Implementation Details}
\label{app:implementation-details}

All probes are trained with AdamW~\cite{adamw} and a linear-warmup cosine learning-rate schedule.
We use batch size 8 for ScanNet semantic tagging and instance grouping, and batch size 10 for DL3DV geometry.
Semantic tagging, instance grouping, and 3D geometry are trained for 10, 40, and 60 epochs, respectively.
Semantic tagging is trained on one NVIDIA A100-80G GPU, while instance grouping and 3D geometry are trained on two NVIDIA A100-80G GPUs.
The tagging probe uses learning rate \(3\times10^{-4}\), weight decay 0.05, two warmup epochs, a depth-2 backbone with width 512, and a two-layer semantic decoder initialized from CLIP ViT-L/14 class-name embeddings.
It is trained with asymmetric multi-label loss with \(\gamma_{\mathrm{neg}}=4\), \(\gamma_{\mathrm{pos}}=0\), and no probability clipping.
The instance probe uses learning rate \(10^{-3}\), weight decay 0.01, two warmup epochs, a depth-2 backbone with width 1024, and a 32-dimensional instance embedding head.
Its contrastive loss samples 2048 valid pixels per step with margin 1.0; at evaluation, embeddings are clustered by HDBSCAN~\cite{mcinnes2017hdbscan} with minimum cluster size 30, minimum samples 5, and PCA reduction to 8 dimensions.
The geometry probe uses learning rate \(10^{-4}\), weight decay 0.05, ten warmup epochs, a depth-4 backbone with width 1024, and DPT heads with channel sizes \([256,512,1024,1024]\).
Point-map, depth, and camera losses are weighted equally.

For feature selection, each model is evaluated with a fixed intermediate layer rather than choosing the best layer per metric.
For VGMs, we use hidden states from the denoising transformer with an empty text prompt; WAN and CogVideoX features use timestep 749, and OpenSora uses its normalized timestep 0.25.
The main 3D runs use VGM layer 20, InternVL3 layer 18, InternVL3.5 layer 22, Qwen2.5-VL layer 21, and Qwen3-VL layer 22.
For ScanNet semantic tagging and instance grouping, we use the corresponding fixed ScanNet runs encoded in the checkpoint names, with the main representative comparison using WAN/CogVideoX layer 18, InternVL3 layer 18, and Qwen3-VL layer 22.
All tasks use the same 76-frame context construction; semantic and instance probes sample 8 frames, while geometry probes sample 4 frames.
VGM features are spatially pooled to a fixed grid when needed: WAN/OpenSora use \(15\times26\), and CogVideoX/Aether use \(15\times22\); VLM features keep their native visual-token grids.

\section{Probe-Depth Ablation Details}
\label{app:probe-depth-ablation}

Table~\ref{tab:probe-depth-ablation-full} provides the full numerical results behind the probe-depth ablation in the main text.
Although the absolute scores vary with probe depth, the relative ordering of representative models remains stable.

\section{Additional Qualitative Example}

\begin{figure}[t]
    \centering
    \includegraphics[width=\columnwidth]{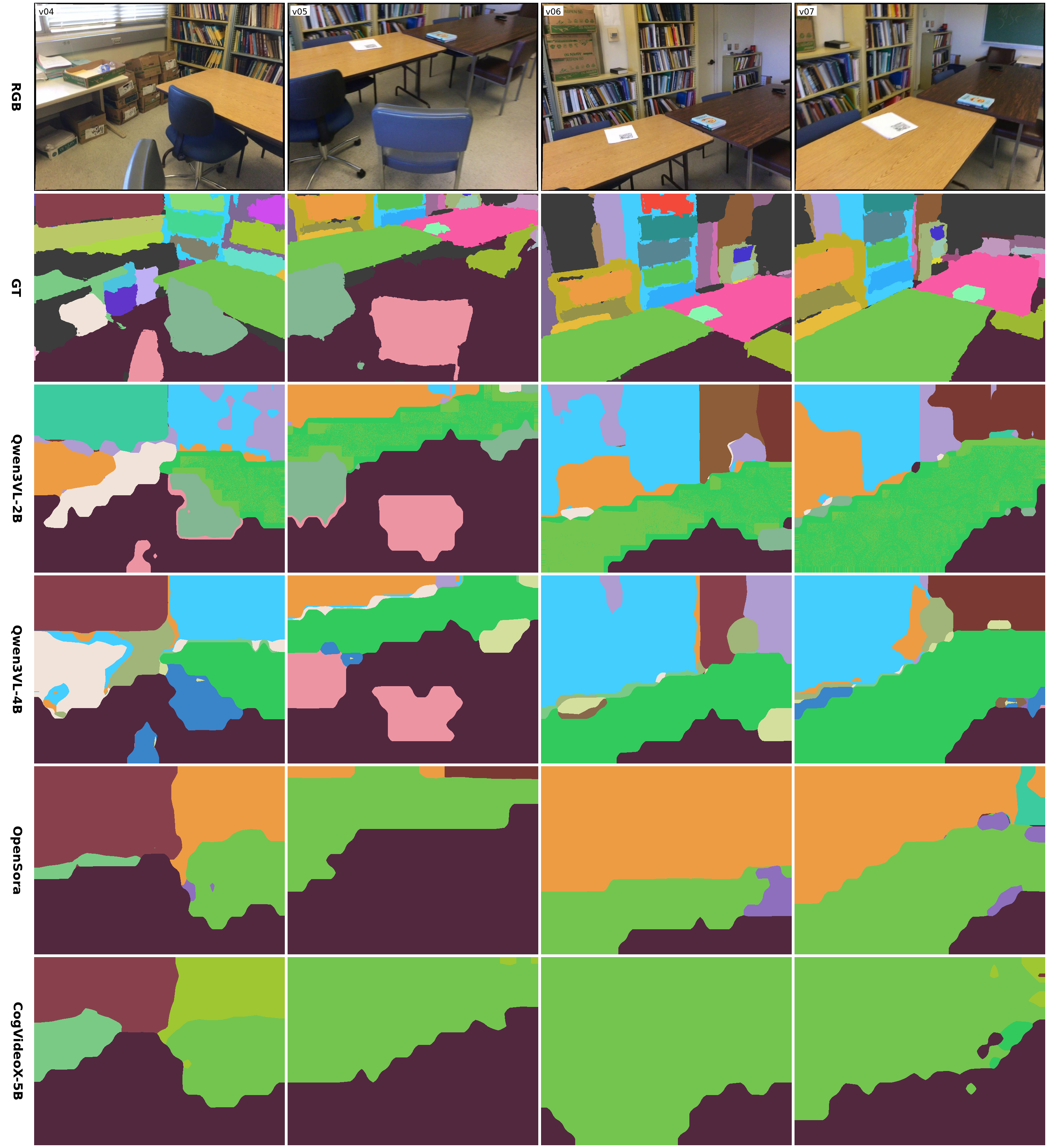}
    \caption{
        Additional instance grouping example on ScanNet scene0030\_01.
        Rows follow Figure~\ref{fig:case-instance-scene0050}: RGB, GT, Qwen3-VL-2B, Qwen3-VL-4B, OpenSora, and CogVideoX-5B.
    }
    \label{fig:case-instance-scene0030}
\end{figure}

Figure~\ref{fig:case-instance-scene0030} shows a cluttered room with tables, chairs, bookshelves, boxes, and small objects observed from multiple views.
The ground truth contains many fine-grained instance regions, especially around the shelves and stacked boxes.
The Qwen3-VL probes are still coarser than the ground truth, but they preserve several meaningful object-level regions such as the tables, chairs, shelves, and foreground objects across views.
In contrast, OpenSora and CogVideoX tend to merge large parts of the scene into only a few broad segments, losing many small objects and object boundaries.

\begin{figure}[t]
    \centering
    \includegraphics[width=\columnwidth]{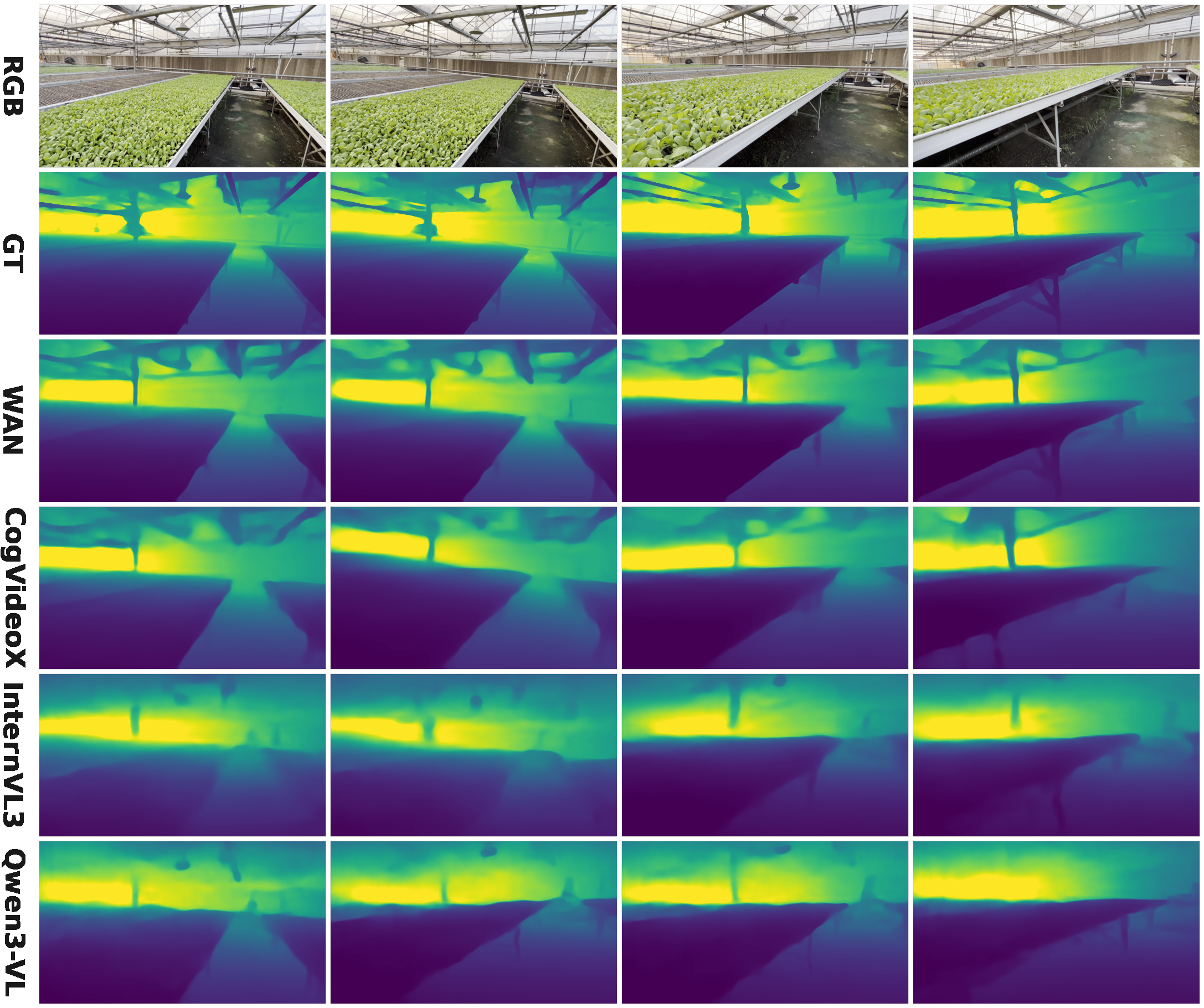}
    \caption{
        Additional depth prediction example on DL3DV.
        Rows show RGB, GT depth, and predictions from WAN, CogVideoX, InternVL3, and Qwen3-VL.
    }
    \label{fig:case-depth-batch0005}
\end{figure}

Figure~\ref{fig:case-depth-batch0005} provides another depth prediction example in a greenhouse scene.
The RGB views contain long planting tables, thin supporting legs, roof beams, and transparent greenhouse structures.
WAN and CogVideoX follow the ground-truth depth more closely on the large table planes and retain sharper discontinuities along table edges and support structures.
InternVL3 and Qwen3-VL recover the coarse near-far layout, but their predictions are visibly smoother and blur several local depth changes, especially near the table boundaries and overhead beams.

Figure~\ref{fig:case-pointcloud-batch0009} visualizes the predicted point maps as colored point clouds for a bookstore scene.
The RGB views show narrow aisles and dense shelves on both sides.
The WAN point cloud keeps a room-like structure close to the GT, with visible shelf planes and a clearer aisle layout.
CogVideoX is noisier but still preserves much of the elongated shelf structure.
By comparison, InternVL3 and Qwen3-VL produce more compact and fragmented point clouds, where the global bookstore layout is harder to read and several shelf planes collapse into less organized geometry.

\end{document}